\documentclass[runningheads]{llncs}
\usepackage[T1]{fontenc}
\usepackage{amsmath,graphicx,hyperref}
\usepackage{cite}
\usepackage{amssymb}
\usepackage{marvosym}
\usepackage{algorithm}
\usepackage{algorithmic}
\usepackage{subfig}
\usepackage{booktabs}
\usepackage{float}
\usepackage{lipsum}
\usepackage{enumitem}
\usepackage{multirow}
\usepackage[table]{xcolor} 
\usepackage{array}
\usepackage{xurl}
\usepackage{orcidlink}
\usepackage{tabularray}
\usepackage{colortbl}
\usepackage{tabularx}

\renewcommand{\orcidID}[1]{\unskip\,\orcidlink{#1}}

\makeatletter
\renewcommand\section{\@startsection{section}{1}{\z@}%
  {-10pt \@plus -2pt \@minus -2pt}%
  {6pt \@plus 2pt}%
  {\normalfont\Large\bfseries}}
\renewcommand\subsection{\@startsection{subsection}{2}{\z@}%
  {-8pt \@plus -2pt \@minus -2pt}%
  {4pt \@plus 2pt}%
  {\normalfont\large\bfseries}}
\makeatother

\setlist{topsep=2pt, partopsep=0pt, parsep=1pt, itemsep=1pt}

\let\OLDthebibliography\thebibliography
\renewcommand\thebibliography[1]{%
  \OLDthebibliography{#1}%
  \setlength{\parskip}{0pt}%
  \setlength{\itemsep}{0pt plus 0.3ex}%
}

\begin{document}

\title{When AI Meets Early Childhood Education: Large Language Models as Assessment Teammates in Chinese Preschools}
\titlerunning{When AI Meets Early Childhood Education}
%
%
\authorrunning{X. Li et al.}
%
%

\author{Xingming Li\inst{1}\orcidID{0009-0002-1927-9980} \and
Runke Huang\inst{2}\textsuperscript{\Letter}\orcidID{0000-0001-9126-8405} \and
Yanan Bao\inst{2}\orcidID{0009-0003-9105-3289} \and
Yuye Jin\inst{2}\orcidID{0009-0008-8115-236X} \and
Yuru Jiao\inst{2}\orcidID{0009-0006-9683-8183} \and
Qingyong Hu\inst{3}\orcidID{0000-0003-0337-9207}}
\authorrunning{X. Li et al.}
\institute{National University of Defense Technology, Changsha, China \\
\email{lixingming@nudt.edu.cn}
\and
The Chinese University of Hong Kong, Shenzhen, China \\
\email{runkehuang@cuhk.edu.cn, \{yananbao,yuyejin\}@link.cuhk.edu.cn, yurujiao03@gmail.com}
\and
University of Oxford, Oxford, United Kingdom \\
\email{huqingyong15@outlook.com}}

\maketitle           
\begin{abstract}
    High-quality teacher-child interaction (TCI) is fundamental to early childhood development, yet traditional expert-based assessment faces a critical scalability challenge. In large systems like China's—serving 36 million children across 250,000+ kindergartens—the cost and time requirements of manual observation make continuous quality monitoring infeasible, relegating assessment to infrequent episodic audits that limit timely intervention and improvement tracking. 
    In this paper, we investigate whether AI can serve as a scalable assessment teammate by extracting structured quality indicators and validating their alignment with human expert judgments.
    Our contributions include: (1) \textbf{TEPE-TCI-370h} (\textit{Tracing Effective Preschool Education}), the first large-scale dataset of naturalistic teacher-child interactions in Chinese preschools (370 hours, 105 classrooms) with standardized ECQRS-EC and SSTEW annotations; (2)  We develop \textit{Interaction2Eval}, a specialized LLM-based framework addressing domain-specific challenges—child speech recognition, Mandarin homophone disambiguation, and rubric-based reasoning—achieving up to 88\% agreement; (3) Deployment validation across 43 classrooms demonstrating an \textbf{18$\times$} efficiency gain in the assessment workflow, highlighting its potential for shifting from annual expert audits to monthly AI-assisted monitoring with targeted human oversight. This work not only demonstrates the technical feasibility of scalable, AI-augmented quality assessment but also lays the foundation for a new paradigm in early childhood education—one where continuous, inclusive, AI-assisted evaluation becomes the engine of systemic improvement and equitable growth. Our project page is available at \url{https://qingyonghu.github.io/Interaction2Eval/}.

    \end{abstract}

\section{Introduction}
\label{sec:intro}

High-quality teacher-child interaction (TCI) is the cornerstone of effective early childhood education (ECE), directly influencing children's cognitive, linguistic, and socio-emotional development \cite{burchinal2005predicting, sylva2004effective, huang2025quality}. Grounded in Vygotsky’s sociocultural theory (1978), learning is understood as a socially mediated process in which language and interaction function as fundamental tools for thinking and development~\cite{vygotsky1978mind}. Especially for preschoolers who are in the early stages of developing abstract thinking and independent learning abilities, interactions with adults and peers remain a primary driver of learning and development~\cite{piaget1952origins}. Current quality evaluation approach of TCI rely on trained observers using standardized instruments like the Childhood Quality Rating Scale-Emergent Curriculum (ECQRS-EC) \cite{sylva2025early} and Sustained Shared Thinking and Emotional Wellbeing (SSTEW) \cite{SSTEW}. However, this approach requires hours of in-person observation and scoring per classroom and is \textit{fundamentally unscalable} and \textit{low efficiency}—a major challenge for large systems like China's, serving nearly 36 million children across over 250,000 kindergartens \cite{moe_preschool_data_2024}. 

Recent advances in automatic speech recognition (ASR) and large language models (LLMs) suggest a promising alternative: \textbf{can we automatically assess teacher-child interaction quality directly from classroom audio?} While this concept seems intuitive, it presents unprecedented technical challenges. Unlike adult speech processing in controlled environments, preschool classrooms feature overlapping child voices, high background noise, non-standard pronunciation, and domain-specific educational terminology \cite{foulkes2010sound}. Moreover, meaningful quality assessment requires more than accurate transcription: it also depends on identifying subtle pedagogical patterns and mapping
them to standardized rubric indicators. Answering this question therefore requires addressing three interrelated issues: \textit{whether large-scale naturalistic classroom data can be collected and annotated with sufficient reliability}, \textit{whether LLM-based systems can achieve reliable agreement with trained human experts on language-accessible rubric indicators}, and \textit{whether such a system can be deployed in real preschool settings with clear operational value}.

\begin{figure}[!t]
\centering
\vspace{-2pt}
\includegraphics[width=\columnwidth]{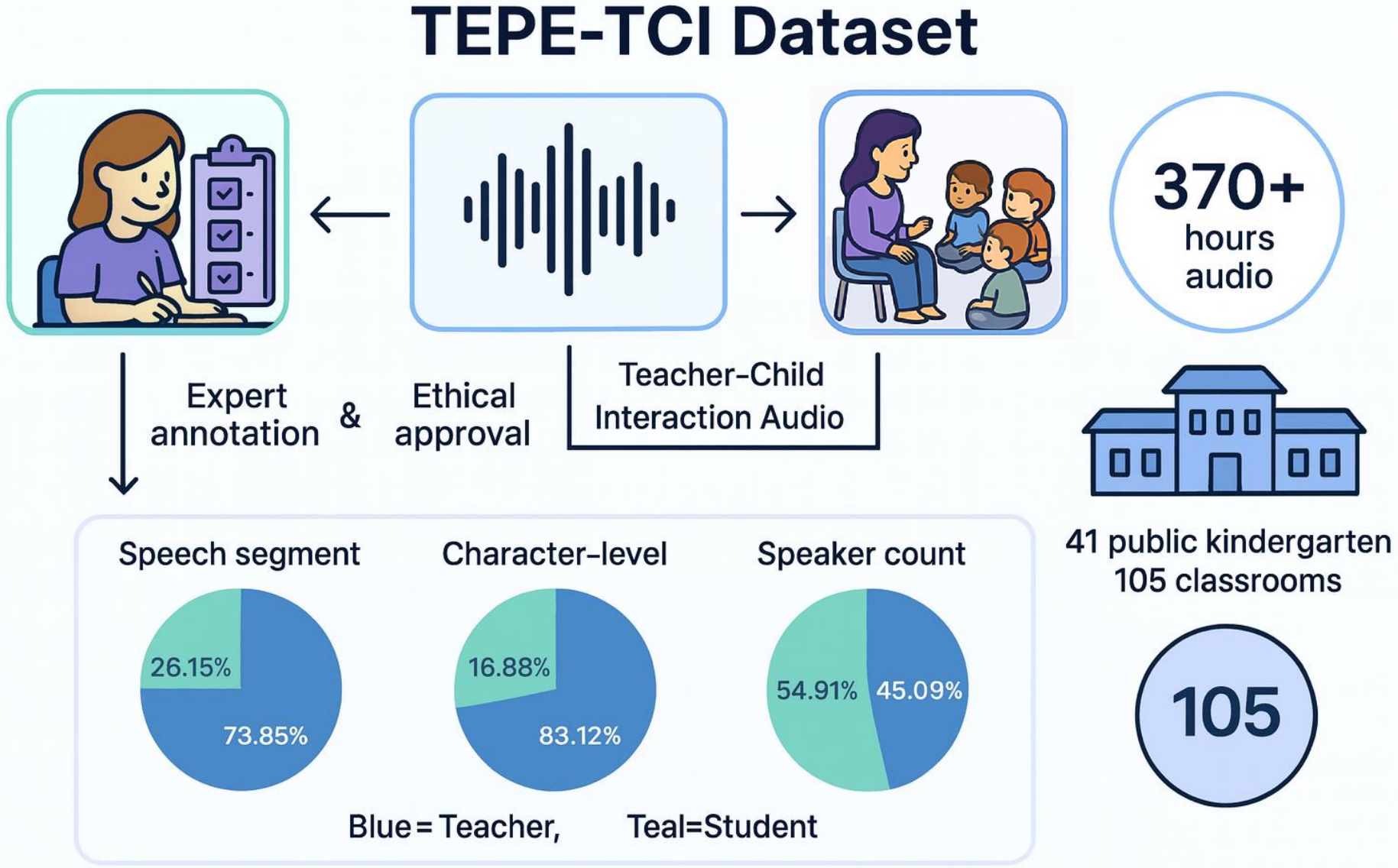}
\vspace{-8pt}
\caption{Overview of the \textbf{TEPE-TCI} dataset, comprising over \textbf{370} hours of teacher–child interaction audio from \textbf{41} public preschools (\textbf{105} classrooms). Data were ethically approved, expert-annotated, and analyzed across levels.}
\label{fig:dataset}
\vspace{-6pt}
\end{figure}

To bridge this gap, we introduce TEPE-TCI, the first comprehensive TCI dataset in Chinese preschools, enabling \textbf{automated teacher-child interaction assessment from classroom audio.}
This represents a new challenge at the intersection of speech processing, natural language processing, and educational measurement.
 Unlike prior work focusing on structured settings like junior or high school dialogues with clear speech and minimal noise \cite{li2025advancing, long2024enhanced}, our task addresses the complexities of naturalistic preschool interactions, which are more \textit{spontaneous, noisy,} and \textit{linguistically diverse}. Additionally, our approach emphasizes \textit{language-mediated interaction aspects}, enabling \textbf{scalable, rubric-grounded assessment} using real-world classroom recordings. Building on this dataset, we develop Interaction2Eval, an automated assessment framework that combines large-scale naturalistic data collection, specialized speech and language processing, and real-world deployment validation. This work demonstrates both the technical feasibility and practical scalability of AI-augmented quality assessment in early childhood education. 
 
We make three primary contributions:

\begin{itemize}[leftmargin=*]
\setlength{\itemsep}{0pt}
\setlength{\parsep}{0pt}
\setlength{\parskip}{0pt}
    \item We present \textbf{TEPE-TCI-370h}, the first comprehensive dataset of naturalistic classroom interactions with expert quality annotations in Chinese preschool contexts, comprising 370 hours of audio from 105 classrooms.
    \item We develop \textbf{Interaction2Eval}, a specialized LLM-based framework addressing domain-specific challenges including child speech recognition, Mandarin homophone disambiguation, and rubric-based reasoning, achieving up to 88\% agreement in interaction quality assessment.
    \item We validate our approach through real-world deployment across 43 classrooms, demonstrating an \textbf{18$\times$} efficiency gain in the assessment workflow and the potential for shifting from annual expert audits to continuous AI-assisted monitoring.
    
\end{itemize}

\section{Related Works}
\label{sec:related}

\noindent \textbf{Educational Speech and Dialogue Datasets.} 
While general speech datasets are abundant, educational speech data for early childhood contexts remains scarce, as summarized in Table~\ref{tab:dataset_comparison}. Most existing children's speech corpora were developed for speech recognition rather than educational assessment, including CSLU Kids \cite{shobaki2007cslu}, CMU Kids \cite{eskenazi1997cmu}, PF-STAR \cite{batliner2005pf_star}, and CHILDES \cite{pye1994childes}. Datasets targeting classroom interactions predominantly focus on K-12 settings. TalkMoves \cite{suresh2022talkmoves} provides 567 mathematics lesson transcripts with discursive move annotations, NCTE \cite{demszky2023ncte} contains elementary classroom recordings with CLASS observation scores, MyST \cite{pradhan2024my} offers 393 hours of conversational speech from grades 3-5 students in tutoring contexts, and SimClass \cite{attia2025simclass} synthesizes classroom audio with ambient noise. For early childhood contexts, resources are notably limited: WSW \cite{sun2025said} captures preschool speech via wearables, Playlogue \cite{kalanadhabhatta2024playlogue} offers 33 hours of adult-child dialogue for naturalistic play, and NCRECE PreK provides pre-kindergarten videos primarily for U.S. research \cite{pianta2016national, pianta2017early}. Notably, these datasets predominantly focus on English-speaking environments and lack standardized quality assessment annotations suitable for teacher-child interaction evaluation. Publicly available children's speech corpora for Chinese are even more limited, particularly for preschool educational contexts. The few existing datasets, including ChildMandarin \cite{zhou2025childmandarin}, SingaKids \cite{chen2016singakids}, and SLT-CSRC \cite{yu2021slt}, were designed for speech recognition tasks and either lack classroom recordings or contain no quality assessment annotations, which restricts their utility for automated interaction assessment. This gap is particularly significant given the unique challenges in Chinese preschool speech processing, including pervasive Mandarin homophone ambiguities, domain-specific educational terminology, and culturally-shaped pedagogical practices.

\begin{table}[tbp]
    \centering
    \vspace{-4pt}
    \caption{Summary of children's speech and educational interaction datasets. K denotes kindergarten, G denotes grade. Diar. indicates speaker diarization and Quality FW indicates quality assessment framework.}
    \label{tab:dataset_comparison}
    \vspace{-2pt}
    \resizebox{\columnwidth}{!}{%
    \begin{tabular}{@{}lcccccccc@{}}
    \toprule
    \textbf{Corpus} & \textbf{Language} & \textbf{Age Range} & \textbf{\# Speakers} & \textbf{Hours} & \textbf{Style} & \textbf{Diar.} & \textbf{Quality FW} & \textbf{Year} \\
    \midrule
    CHIEDE \cite{garrote2008chiede} & Spanish & 3-6 & 59 & $\sim$8 & Conversation & Partial & - & 2008 \\
    CSLU Kid's Speech Corpus~\cite{shobaki2007cslu} & English & K-G10 & 1,100 & - & Read+Spont. & N & - & 2007 \\
    CMU Kids Corpus~\cite{eskenazi1997cmu} & English & 6-11 & 76 & - & Read speech & N & - & 1997 \\
    PF-STAR Children’s Speech \cite{batliner2005pf_star} & English & 4-14 & 158 & 14.5 & Read Speech & N & - & 2005 \\
    MyST Corpus~\cite{pradhan2024my} & English & G3-G5 & 1,371 & 393 & Conversation & Y & - & 2024 \\
    TalkMoves \cite{suresh2022talkmoves} & English & K-12 & - & - & Classroom & Y & - & 2022 \\
    NCTE \cite{demszky2023ncte} & English & G4-G5 & 317 & 1,660 les. & Classroom & Y & CLASS+MQI & 2023 \\
    SimClass \cite{attia2025simclass} & English & - & - & 391 & Simulated & N & - & 2025 \\
    WSW \cite{sun2025said} & English & 3-5 & 17 & 1,592 & Classroom & Y & - & 2025 \\
    Playlogue \cite{kalanadhabhatta2024playlogue} & English & Preschool & - & 33 & Play-based & Y & DPICS & 2024 \\
    \midrule
    SingaKids \cite{chen2016singakids} & Chinese & 7-12 & 255 & 75 & Reading & N & - & 2016 \\
    SLT-CSRC C1 \cite{yu2021slt} & Chinese & 7-11 & 927 & 28.6 & Reading & N & - & 2021 \\
    SLT-CSRC C2 \cite{yu2021slt} & Chinese & 4-11 & 54 & 29.5 & Conversation & N & - & 2021 \\
    ChildMandarin \cite{zhou2025childmandarin} & Chinese & 3-5 & 397 & 41.3 & Conversation & N & - & 2024 \\
    \midrule
    \textbf{TEPE-TCI (Ours)} & \textbf{Chinese} & \textbf{3-4} & \textbf{2,550} & \textbf{370} & \textbf{Classroom} & \textbf{Y} & \textbf{ECQRS-EC+SSTEW} & \textbf{2025} \\
    \bottomrule
    \end{tabular}%
    }
    \vspace{-6pt}
    \end{table}

\noindent \textbf{LLMs for Educational Assessment.} Large language models have shown substantial progress in educational assessment, with GPT-4 achieving human-comparable grading accuracy \cite{mendoncca2025evaluating, impey2025grade} and high inter-coder agreement in classroom dialogue analysis \cite{long2024evaluating, li2025advancing}. For early childhood education, Wang \textit{et al.} \cite{wang2025classroom} explored LLM-based instructional support evaluation using CLASS protocols \cite{CLASS}, while Whitehill \textit{et al.} \cite{whitehill2023automated} developed automated the CLASS scoring with utterance-level feedback. However, these approaches focus on a single domain of CLASS framework (e.g., instructional support), rather than capturing the full range of interaction quality. Recent work has explored LLMs for developmental assessment~\cite{yang2025validating}, yet comprehensive teacher-child interaction assessment using multi-dimensional professional scales remains unexplored, particularly in non-English contexts where cultural and linguistic factors create distinct challenges.

To our knowledge, no existing system addresses automated ECQRS-EC or SSTEW assessment, nor audio-based teacher-child interaction evaluation in Chinese preschools that spans multiple classroom scenarios. Our work establishes both a new benchmark dataset and initial baselines for this task.

\section{Problem Formulation}
\label{sec:problem}

We formalize \textbf{automated teacher-child interaction assessment in preschool environments} as follows: Given a noisy, multi-speaker classroom audio recording $A$ of duration $T$, the goal is to detect the presence or absence of behavioral indicators defined in standardized educational rubrics such as ECQRS-EC and SSTEW~\cite{sylva2025early,SSTEW}. These rubrics are widely used in ECE to assess the TCI quality based on the occurrence of developmentally supportive teaching behaviors. Each rubric item comprises multiple indicators describing concrete, observable teacher behaviors
associated with varying levels of quality - from inadequate to excellent. During 3-hour
classroom observation, observers assess whether such behaviors occur and assign scores
accordingly. For each indicator i, the system outputs a binary judgment $y_i \in \{0, 1\}$ indicating whether the corresponding behavior was observed in the interaction. Item-level scores are subsequently derived from indicator patterns following official scoring protocols.

\subsection{Technical Challenges}

This task introduces several challenges unique to the preschool-based speech processing and educational assessment:

\textbf{Data Challenges:} As reviewed in Section~\ref{sec:related}, publicly available children's speech corpora for Chinese preschool contexts are extremely limited. Existing datasets either lack naturalistic classroom recordings or contain no standardized quality assessment annotations, making it impossible to train or evaluate automated interaction assessment systems for this domain. 

\textbf{Speech Processing Challenges:} Chinese preschool classrooms present compounded acoustic and linguistic difficulties. Acoustically, recordings feature high background noise, overlapping multi-speaker speech, non-standard child pronunciation, and distant-microphone conditions. Linguistically, Mandarin's tonal nature causes extensive homophone ambiguity (\textit{e.g.}, \textit{chénfú}: [sinking/floating] vs. [submission]), while domain-specific educational terminology (\textit{e.g.}, \textit{jìnqū}: [enter learning centers] vs. [go inside]) and teachers' child-directed speech patterns further challenge standard ASR systems. 

\textbf{Assessment Challenges:} Moving from transcription to meaningful evaluation requires handling \textit{rubric complexity}—educational assessment scales contain nuanced criteria requiring deep contextual understanding. \textit{Temporal reasoning} is essential as quality judgments must consider interaction patterns across extended time periods. Additionally, \textit{pedagogical knowledge} is also crucial for accurate assessment, requiring understanding of early childhood education principles and developmental appropriateness.

\subsection{Task Scope and Limitations}

We focus on \textbf{language-accessible aspects} of teacher-child interaction—those dimensions that can be reliably evaluated from verbal exchanges alone. While we acknowledge that high-quality interaction encompasses non-verbal elements (gestures, spatial arrangement, materials), our approach addresses the substantial subset of assessment criteria that depend on conversational patterns, questioning strategies, and linguistic scaffolding.

\section{Dataset Construction}
\label{sec:dataset}

\subsection{Data Collection Protocol}

We collected audio using professional recording equipment (iFLYTEK H1 Pro) to capture naturalistic teacher-child interactions across diverse classroom contexts including group activities, free play, outdoor activities, and daily routines.

\textbf{Scale and Scope:} Forty-one preschools participated in this study, spanning three quality tiers as defined by the local education authority: district-level (N=14), municipal-level (N=12), and provincial-level (N=15). These tiers reflect differences in overall institutional quality and operating conditions. From each preschool, two to three K1 classrooms serving 3 to 4 years old were recruited, with each classroom accommodating approximately 25 students. Eight research assistants conducted data collection over 6 weeks, resulting in 370+ hours of audio from 105 classrooms with average session length of approximately 3.5 hours.

\textbf{Ethical Consideration:} This study was approved by the University Ethics Committee of The Chinese University of Hong Kong, Shenzhen (Approval No. EF20241026001). Informed consent was obtained from all teachers and from parents or legal guardians, who were fully briefed on the study's purpose, recording procedures, data handling protocols, and intended use for academic research. Parents or legal guardians retained the right to withdraw their child's data at any time without penalty. To minimize privacy risks, speaker diarization preserves only role labels (teacher/child), and all recordings, transcripts, and annotations are securely stored with access restricted to authorized researchers. For future data sharing, we plan to release only anonymized transcripts, expert annotations, and supporting documentation for non-commercial academic research. Because classroom audio from young children may contain identifiable voice information, raw audio recordings will not be publicly released; any exceptional access would require additional ethical review and formal data-use agreements.

\subsection{Expert Annotation Process}
Professional experts (the same 8 assessors) evaluated the quality of the teacher-child interaction.

\textbf{Annotation Protocol:} Quality assessment followed standardized ECQRS-EC
    and SSTEW protocols. ECQRS-EC comprises 22 items spanning four domains: Literacy, Mathematics, Science \& Environment, and Diversity. SSTEW includes 15 items covering areas such as building independence and emotional wellbeing, language support, and critical thinking. Each item is organized into four performance levels (1=inadequate, 3=minimal, 5=good, 7=excellent), operationalized through behavioral indicators at each level (\textit{e.g.}, Level 3: ``Children are allowed to talk among themselves''; Level 7: ``Adults scaffold children's conversations''). Our annotation adopted indicator-level binary coding: assessors judged whether each specified behavior was observed (1) or not (0). Item-level scores were subsequently derived from indicator patterns following official scoring protocols: the score was assigned at the highest level for which all indicators were met; if more than 50\% of the indicators for the next level were also satisfied, the midpoint between those levels was assigned.
    
    Our study prioritized indicators that are most representative of daily classroom practice and feasible for audio-based observation. As a result, our annotations covered 17 of the 22 ECQRS-EC items and 14 of the 15 SSTEW items, comprising a total of 112 ECQRS-EC indicators and 94 SSTEW indicators.

\textbf{Quality Assurance:} All assessors underwent extensive training on both assessment scales, with inter-rater reliability validation requiring $\kappa > 0.80$ agreement before independent scoring. During the validation phase, two assessors scored each classroom; after achieving reliability thresholds, single assessors completed evaluations. This rigorous process ensured high-quality ground truth annotations essential for automated system development.

\subsection{Dataset Characteristics and Processing}

Figure \ref{fig:dataset} presents comprehensive statistics of the  \textbf{TEPE-TCI-370h} dataset. Speech segment analysis reveals teachers contributing 73.85\% of segments while students account for 26.15\%. Character-level analysis shows even greater teacher dominance (83.12\% vs 16.88\%), reflecting typical classroom discourse patterns where teachers provide more extended explanations and instructions. Interestingly, speaker count analysis shows more balanced participation (45.09\% teachers vs 54.91\% students), indicating active child engagement despite shorter individual contributions. This dataset fills a critical gap as the first large-scale Chinese preschool resource combining naturalistic classroom audio with standardized quality annotations.

\section{Interaction2Eval Framework}
\label{sec:method}

\begin{figure}[thb]
    \centering
    \vspace{-2pt}
    \includegraphics[width=0.9\columnwidth]{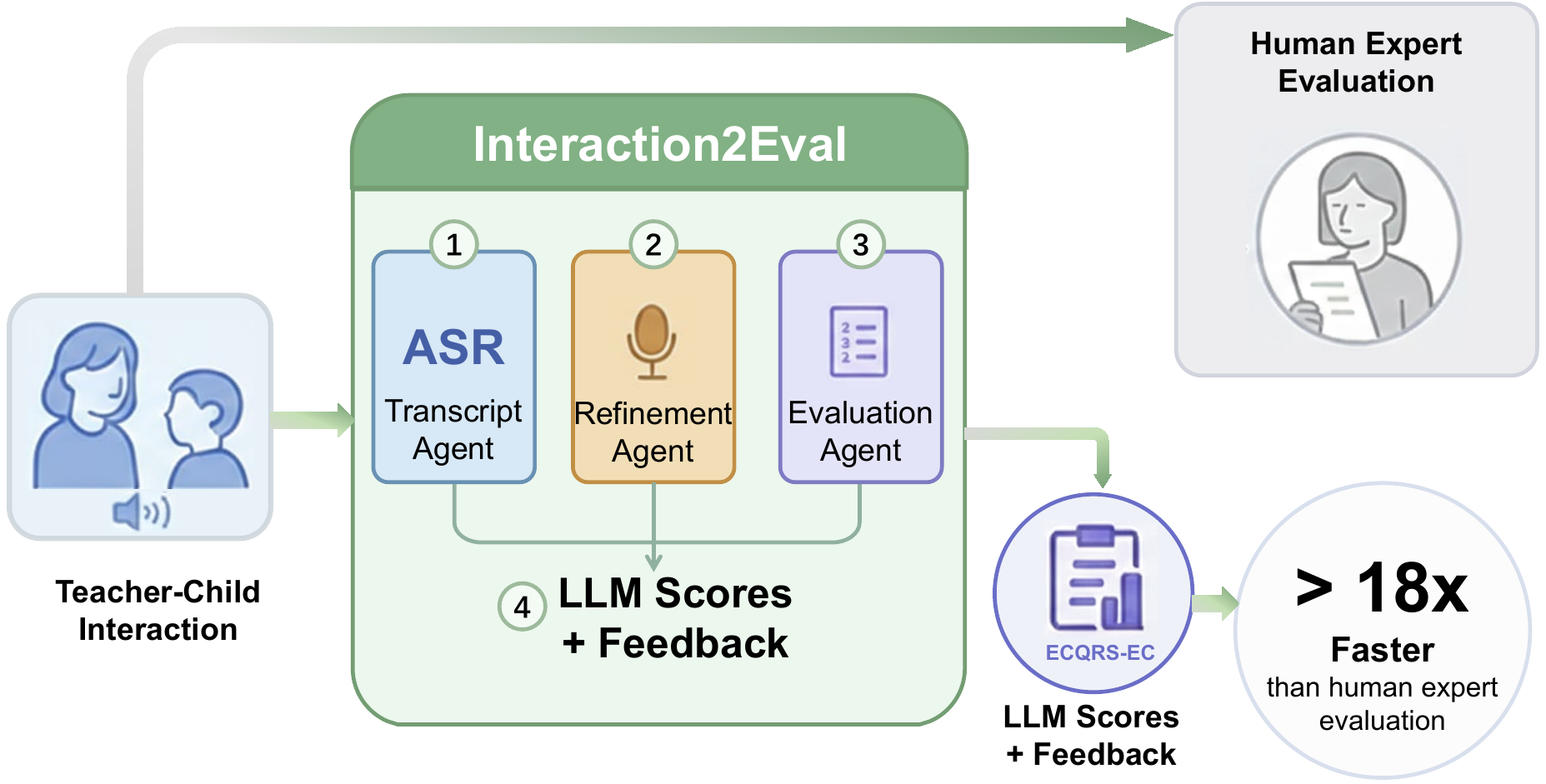}
    \vspace{-6pt}
    \caption{Overview of the \textit{Interaction2Eval} pipeline.}
    \label{fig:pipeline}
    \vspace{-4pt}
    \end{figure}

Building on insights from dataset construction, we develop \textit{Interaction2Eval} (Figure~\ref{fig:pipeline}), a specialized framework addressing core challenges of automated assessment through three LLM-empowered agents (\textit{Transcription}, \textit{Refinement}, and \textit{Evaluation} Agent)

\subsection{Transcription and Refinement Agents}

The first stage transforms raw audio into assessment-ready transcripts. Initial ASR processing using Paraformer with diarization and punctuation restoration \cite{gao2023funasr,bredin2023pyannote} revealed systematic domain-specific errors: professional transcribers categorized errors as homophones (51.67\%), extra words (20.80\%), speaker identification (13.72\%), punctuation/segmentation (7.75\%), and omissions (6.06\%). The predominance of homophone errors—significantly higher than general speech tasks—motivated our Refinement Agent design.

Given this challenge, the \textit{Refinement Agent} leverages large language models for context-aware correction. The agent prompt incorporates educational domain knowledge, explicitly stating that the input is preschool classroom speech likely containing homophone errors common in educational settings. To guide disambiguation, the prompt includes examples of frequent confusion pairs—such as \textit{jìnqū} (enter learning centers) versus \textit{jìnqù} (go inside), and \textit{chénfú} (sinking/floating) versus \textit{chénfú} (submission)—and instructs the model to correct errors while preserving original meaning and speaker attributions.
To accommodate context length limits, processing proceeds via sliding window: the model receives transcript segments, produces corrected versions, and corrections are realigned with original timestamps and speaker labels.

\subsection{Rubric-Based Evaluation Agent}

The \textit{Evaluation Agent} automates the assessment of teacher-child interactions by detecting behavioral indicators defined in the \textit{SSTEW} and \textit{ECQRS-EC} rubrics. We focus specifically on language-accessible indicators—those that can be reliably identified from verbal exchanges, such as whether teachers use open-ended questions or encourage child-initiated dialogue.

To construct the agent, we collaborated with expert assessors to translate rubric specifications into structured prompts. Each prompt provides the model with indicator definitions across performance levels, accompanied by contrastive examples that illustrate both high-quality and low-quality interaction patterns. The model follows an evidence-first reasoning process: it first locates relevant utterances in the transcript, then determines whether the target indicator is present, and finally provides justification grounded in specific textual evidence.

For each indicator, the agent outputs a binary judgment indicating presence or absence, along with the supporting transcript segment. The agent also generates pedagogical suggestions to support teacher professional development. Through iterative prompt refinement with domain experts, we addressed common failure modes including hallucinated evidence and misalignment between detected behaviors and rubric definitions.

\section{Experiments and Results}

\subsection{Transcription Quality Evaluation}

To ensure high-quality transcriptions, we compared two leading ASR models, Whisper-large-v3~\cite{radford2023robust} and FunASR~\cite{gao2023funasr}, on a 5-hour test set with 16,168 reference characters. Performance was measured using \textit{Character Error Rate (CER)}. As shown in Table~\ref{tab:wer}, raw transcription errors were significant due to homophones, overlapping speech, and domain-specific vocabulary. Whisper-large-v3 exhibited a high raw CER of 35.1\%, likely due to suboptimal Mandarin adaptation, while FunASR Paraformer achieved a lower initial CER of 9.9\%. Our \textit{Refinement Agent}, powered by the Qwen3-Max \cite{yang2025qwen3}, reduced errors by addressing homophone ambiguities, lowering Paraformer's CER to \textbf{4.3\%} (\textbf{56.6\% relative improvement}) and Whisper's to 23.2\%. These results demonstrate the agent's effectiveness, particularly for Mandarin-specific challenges.

\begin{table}[!t]
    \centering
    \vspace{-0.35cm}
    \caption{\textit{CER} results. $\Delta$ / $\downarrow$ : absolute/relative reduction.}
    \label{tab:wer}
    \scalebox{0.75}{
    \begin{tabular}{@{}lccc@{}}
    \toprule
    Model & Raw CER (\%) & After Refinement (\%) & $\Delta$ (\%) $\downarrow$ \\
    \midrule
    Whisper-large \cite{radford2023robust}   & 35.1 & 23.2 & -11.9 ($\downarrow$33.4\%) \\
    FunASR Paraformer \cite{gao2023funasr} & 9.9  & 4.3  & -5.6  ($\downarrow$56.6\%) \\
    \bottomrule
    \end{tabular}}
    \vspace{-0.2cm}
    \end{table}

\subsection{Error Analysis and Domain Adaptation}

\begin{figure}[h]
\centering
\vspace{-4pt}
\includegraphics[width=0.8\columnwidth]{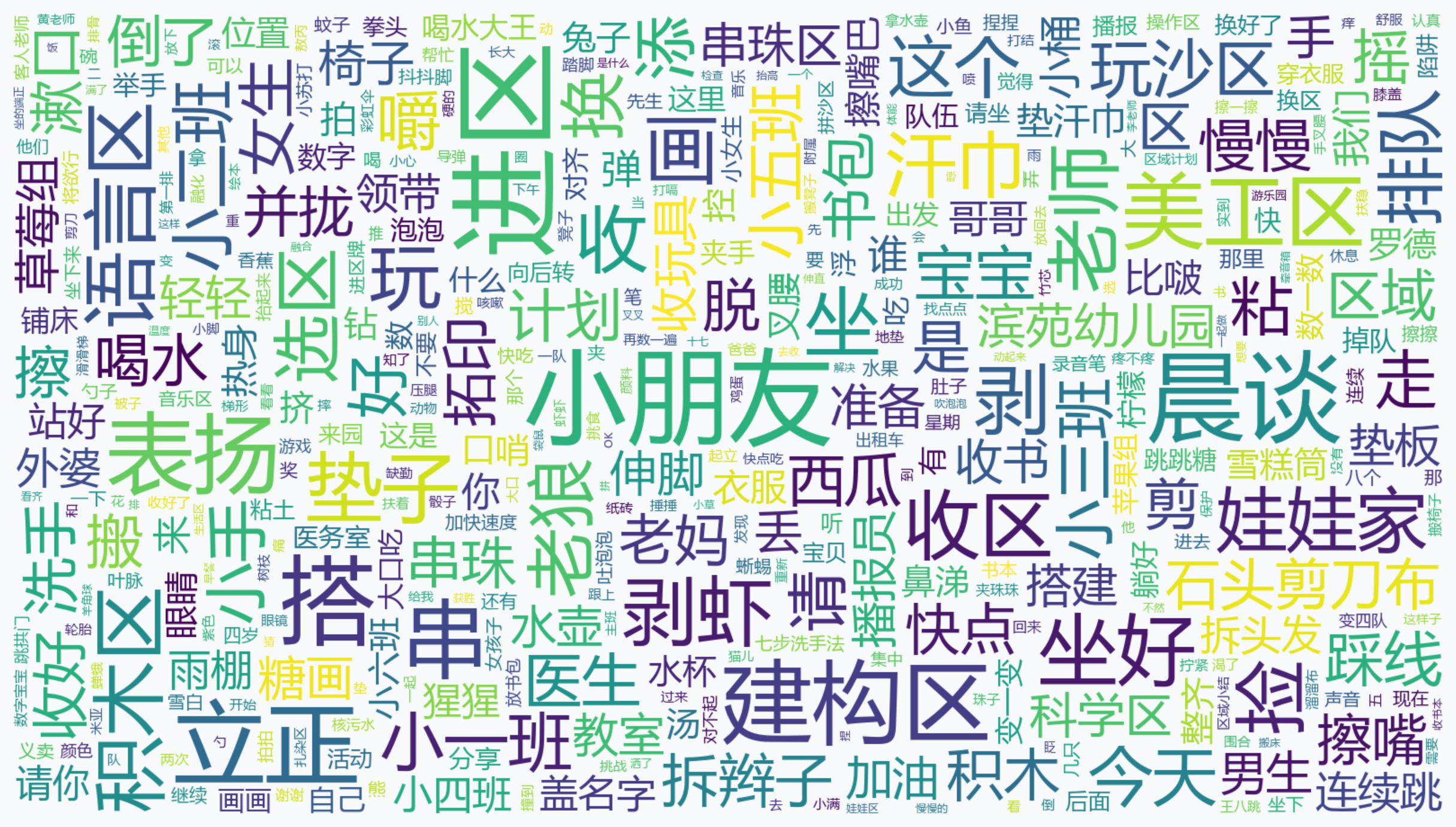}
\vspace{-8pt}
\caption{Frequently misrecognized terms in raw ASR outputs (term size reflects error frequency).}
\label{fig:wordcloud}
\vspace{-6pt}
\end{figure}

Figure \ref{fig:wordcloud} visualizes the most frequently misrecognized terms in raw ASR outputs, with term size reflecting error frequency. These errors concentrate heavily in education-specific vocabulary commonly used in teacher-child interactions, including homophonic pairs like (jìn qū: enter learning centers) vs (jìn qù: go inside). This analysis confirms that semantic-level language modeling is critical for accurate transcription in specialized educational settings and validates the importance of our domain-aware refinement approach.

\subsection{Assessment Consistency Analysis}

To evaluate the robustness of automated scoring across diverse LLM architectures and cultural alignments, we benchmark four state-of-the-art models: two international (GPT-5~\cite{gpt5_system_card}, Gemini-2.5-pro~\cite{comanici2025gemini}) and two Chinese-optimized (DeepSeek-v3.1~\cite{liu2024deepseek}, Qwen3-Max~\cite{yang2025qwen3}). Results also reveal three key findings:

\textbf{(1) Chinese-adapted LLMs outperform international counterparts}, particularly on both ECQRS-EC and SSTEW dimensions. DeepSeek-v3.1 achieves the highest mean agreement on both scales (87.3\% for ECQRS-EC, 87.9\% for SSTEW), followed by Qwen3-Max (85.7\% and 86.6\% respectively). This advantage likely stems from their alignment with Mandarin linguistic patterns, pedagogical terminology, and local classroom discourse norms, highlighting the critical role of \textit{cultural and linguistic grounding} in educational AI systems.

\textbf{(2) Performance on SSTEW consistently exceeds that on ECQRS-EC} across almost all LLMs. This may be attributed to the distinct focuses of the two rubrics: ECQRS-EC focuses on content delivery (\textit{e.g.,} language, early math, science, literacy) while SSTEW indicators focus on how teachers think and talk with children. Therefore, ECQRS-EC requires a more intent-sensitive and curriculum-aligned form of inference, while SSTEW can often be scored based on observable language structure, affective cues, and scaffolding patterns — areas where language models naturally excel.

\textbf{(3) Performance is promising but imperfect—leaving room for future work.}  While top models reach 87.9\% agreement (DeepSeek-v3.1 on SSTEW), $\kappa$ values of 0.71--0.74 indicate moderate-to-substantial but not yet expert-level agreement. For reference, inter-rater reliability among trained human experts in our annotation process was $\kappa \approx 0.82-0.85$, suggesting that the best-performing LLMs achieve approximately 85–90\% of human expert consistency. This gap highlights the challenge of encoding complex rubric logic (\textit{e.g.,} "sustained shared thinking") into prompts, and underscores TEPE-TCI's role as a \textit{benchmark for improvement}, not a final solution.

These findings validate the feasibility of audio-driven automated assessment, while clearly delineating its current limits—especially for context-heavy constructs like ECQRS-EC.

\begin{table*}[!t]
    \centering
    \vspace{-4pt}
   \caption{Indicator-level agreement between LLM predictions and human expert annotations for ECQRS-EC \cite{sylva2025early} and SSTEW \cite{SSTEW} scales, measured by percentage agreement (\%Agr.) and Cohen's Kappa ($\kappa$) which quantifies true agreement beyond chance following \cite{cohen1960coefficient}.}
    \label{tab:eval_consistency}
    \vspace{-2pt}
    \renewcommand{\arraystretch}{1.1}
    \setlength{\tabcolsep}{4pt}
    \resizebox{\textwidth}{!}{%
    \begin{tabular}{@{}llcccccccc@{}}
    \toprule
    & & \multicolumn{8}{c}{\textbf{Models}} \\
    \cmidrule(l){3-10}
    & & \multicolumn{2}{c}{\textbf{GPT-5} \cite{gpt5_system_card}} & \multicolumn{2}{c}{\textbf{Gemini-2.5-pro} \cite{comanici2025gemini}} & \multicolumn{2}{c}{\textbf{DeepSeek-v3.1} \cite{liu2024deepseek}} & \multicolumn{2}{c}{\textbf{Qwen3-Max} \cite{yang2025qwen3}} \\
    \cmidrule(lr){3-4} \cmidrule(lr){5-6} \cmidrule(lr){7-8} \cmidrule(l){9-10}
    \textbf{Scale} & \textbf{Dimension} & $\kappa$ & \%Agr. & $\kappa$ & \%Agr. & $\kappa$ & \%Agr. & $\kappa$ & \%Agr. \\
    \midrule
    \multirow{4}{*}{\textbf{ECQRS-EC}} 
    & Literacy              & 0.678 & 0.844 & 0.736 & 0.871 & 0.705 & 0.886 & 0.764 & 0.882 \\
    & Mathematics           & 0.631 & 0.836 & 0.659 & 0.845 & 0.721 & 0.876 & 0.656 & 0.853 \\
    & Science               & 0.605 & 0.823 & 0.591 & 0.852 & 0.704 & 0.858 & 0.611 & 0.836 \\
    \cmidrule(l){2-10}
    & \textit{Overall Mean} & \textbf{0.638} & \textbf{0.834} & \textbf{0.662} & \textbf{0.856} & \textbf{0.710} & \textbf{0.873} & \textbf{0.677} & \textbf{0.857} \\
    \midrule
    \multirow{5}{*}{\textbf{SSTEW}} 
    & Trust \& Self-regulation    & 0.695 & 0.859 & 0.668 & 0.848 & 0.782 & 0.895 & 0.763 & 0.903 \\
    & Language \& Communication   & 0.718 & 0.863 & 0.752 & 0.878 & 0.867 & 0.949 & 0.825 & 0.913 \\
    & Learning \& Critical Think. & 0.704 & 0.820 & 0.651 & 0.804 & 0.648 & 0.828 & 0.627 & 0.815 \\
    & Planning \& Assessment      & 0.679 & 0.837 & 0.693 & 0.834 & 0.665 & 0.843 & 0.650 & 0.831 \\
    \cmidrule(l){2-10}
    & \textit{Overall Mean} & \textbf{0.699} & \textbf{0.845} & \textbf{0.691} & \textbf{0.841} & \textbf{0.741} & \textbf{0.879} & \textbf{0.716} & \textbf{0.866} \\
    \bottomrule
    \end{tabular}%
    }
    \vspace{-6pt}
\end{table*}

\subsection{Scalability and Efficiency Impact}

Our framework effectively reduces assessment time, marking an important step toward scalable deployment. 
Traditional manual assessment requires approximately 380 minutes of assessment workflow per classroom (240 min in-person observation, 20 min indicator coding, 120 min report writing), with expert presence making evaluation inherently sequential. In contrast, Interaction2Eval completes the corresponding workflow in approximately 21 minutes (5 min audio processing, 12 min transcription and refinement, 4 min evaluation and report generation), achieving an 18$\times$ efficiency gain. This comparison reflects the time required to complete the assessment procedure for a classroom session. This efficiency gain stems from two factors: (1)  \textit{asynchronous recording} eliminates synchronous observation requirements—teachers record naturally without external observers, enabling parallel assessment at near-zero marginal cost; and (2) \textit{automated processing} reduces manual review through high-accuracy LLM-based refinement (CER 4.3\%), requiring expert validation only for flagged ambiguities than full proofreading. At system scale, this translates to substantial resource savings: assessing 100 classrooms monthly would require 633 expert-hours under traditional protocols (100 $\times$ 380 min), compared to only 35 hours with our framework (100 $\times$ 21 min), enabling the critical shift from annual audits to continuous monitoring across China's 250,000+ kindergarten system where manual assessment remains logistically infeasible.

\begin{figure}[h]
    \centering
    \vspace{-4pt}
    \includegraphics[width=0.8\columnwidth]{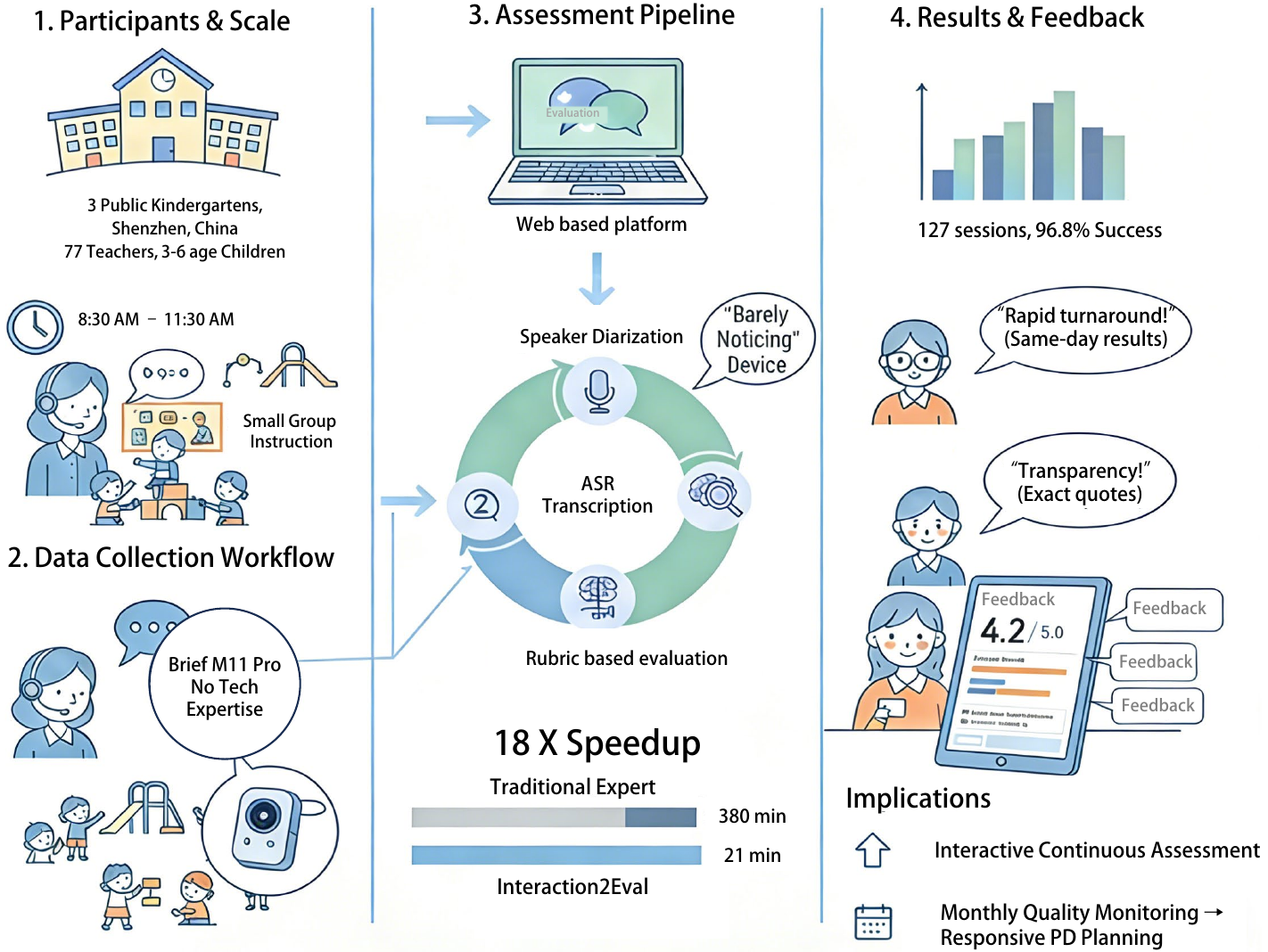}
    \vspace{-8pt}
    \caption{Overview of the pilot deployment study and key outcomes.}
    \label{fig:feedback}
    \vspace{-6pt}
    \end{figure}

\section{From Research to Practice: Deployment Experience}

To evaluate real-world viability, we conducted a pilot deployment of Interaction2Eval in operational preschool settings, examining both system performance and human-AI workflow integration (Figure~\ref{fig:feedback}).

\textbf{Participants and Scale.} We partnered with 3 public kindergartens in Shenzhen, China, covering 43 classrooms serving 3--6 year-old children. Each classroom had one head teacher, one assistant teacher and 25--35 students. Participating teachers (N=77) received brief training on recording equipment usage but required no technical expertise in AI or assessment protocols.

\textbf{Data Collection Workflow.} Teachers wore unobtrusive recording devices (iFLYTEK H1 Pro) during regular morning sessions (8:30--11:30 AM), capturing naturalistic interactions across activities including circle time, free play, and small group instruction. The wireless design ensured minimal disruption to classroom routines, with teachers reporting ``barely noticing'' the device after initial adaptation (typically 2--3 days).

\textbf{Assessment Pipeline.} Upon session completion, teachers uploaded audio files to our web-based platform via one-click transfer. The system automatically executed the full pipeline: (1) speaker diarization, (2) ASR transcription, (3) LLM-based refinement, and (4) rubric-based evaluation with indicator-level feedback. The corresponding assessment workflow required approximately 21 minutes for a 3-hour session, compared to 380 minutes for traditional expert observation and scoring.

\textbf{Results.} Over 4 weeks of preliminary deployment, the system processed 127 classroom sessions with a 96.8\% success rate (4 sessions required manual intervention due to recording quality issues). Average assessment workflow time was 21 minutes per 3-hour session, compared to 380 minutes for traditional manual assessment—an \textbf{18$\times$ efficiency gain in the assessment workflow}. This speedup stems from: (1) asynchronous recording eliminating real-time observation requirements, and (2) automated transcription and scoring reducing manual coding time from hours to minutes.

\textbf{Qualitative Feedback.} Post-deployment interviews with 12 teachers and 3 administrators revealed positive reception. Teachers highlighted rapid turnaround and transparency, noting that results were available same-day rather than weeks later and that exact quotes supporting each score made evaluations understandable. A veteran teacher with 22 years of experience described the indicator-level feedback as a ``data mirror'' that enabled moving from vague intuitions to evidence-based reflection on specific strengths and gaps. Administrators noted the system enabled monthly rather than annual quality monitoring, facilitating more responsive professional development planning.

\textbf{Limitations Observed.} Users correctly identified system boundaries: effective for language-accessible interaction dimensions (dialogue quality, questioning strategies) but unable to assess physical environment or non-verbal engagement—areas requiring multimodal observation. This aligns with our design scope and suggests clear division of labor: AI-assisted continuous monitoring of conversational interactions, complemented by periodic expert evaluation of comprehensive quality including physical and visual aspects.

\textbf{Implications.} These preliminary results validate technical feasibility for scaled deployment. The 18$\times$ efficiency gain in the assessment workflow is not merely quantitative—it enables qualitative transformation from episodic external audits to integrated continuous assessment, supporting the iterative improvement cycles emphasized in contemporary early childhood education quality frameworks.

\section{Conclusion and Future Directions}

We introduce automated teacher-child interaction assessment from audio as a new research area, contributing  \textbf{TEPE-TCI-370h}, the first large-scale preschool interaction dataset in China, alongside Interaction2Eval, a specialized LLM-based framework for this challenging problem. Our work demonstrates that high-quality automated assessment is achievable despite significant technical challenges, opening pathways for scalable ECE quality improvement.

Key findings include the critical importance of domain-specific solutions for educational speech challenges (homophones, terminology), the feasibility of expert-level LLM assessment when properly guided by educational rubrics, and the practical scalability of audio-based assessment while maintaining pedagogical validity. Future research directions include multimodal integration (audio-visual), real-time formative feedback, cross-linguistic generalization, and longitudinal studies on educational quality improvement.

\vspace{6pt}
\noindent\textbf{\ackname} This work was supported by the National Natural Science Foundation of China under Grant No.~62407037 and 62306331.

\bibliographystyle{splncs04}
\bibliography{refs}

\end{document}